%% file: main.tex
\documentclass[10pt,twocolumn,letterpaper]{article}

\usepackage{cvpr}              
\usepackage[accsupp]{axessibility}
\usepackage[hang,flushmargin]{footmisc}

\input{preamble}

\definecolor{cvprblue}{rgb}{0.21,0.49,0.74}
\usepackage[pagebackref,colorlinks,citecolor=cvprblue]{hyperref}

\def\paperID{8265} 
\def\confName{CVPR}
\def\confYear{2024}

\title{Generating Human Motion in 3D Scenes from Text Descriptions}

\author{
    Zhi Cen$^{1}$
    \quad Huaijin Pi$^{1}$
    \quad Sida Peng$^{1}$$^\dagger$
    \quad Zehong Shen$^{1}$
    \\[1mm]
    \quad Minghui Yang$^{2}$
    \quad Shuai Zhu$^{2}$
    \quad Hujun Bao$^{1}$
    \quad Xiaowei Zhou$^{1}$
    \\[2mm]
    $^{1}$Zhejiang University
    \quad $^{2}$Ant Group
}

\begin{document}

\twocolumn[\maketitle\input{figures/figure_teaser}\bigbreak]

\let\thefootnote\relax\footnotetext{The authors from Zhejiang University are affiliated with the State Key Lab of CAD\&CG. $^\dagger$Corresponding author.}
\input{sec/0_abstract}
\input{sec/1_intro}
\begin{figure*}
    \centering
    \includegraphics[width=0.9\linewidth]{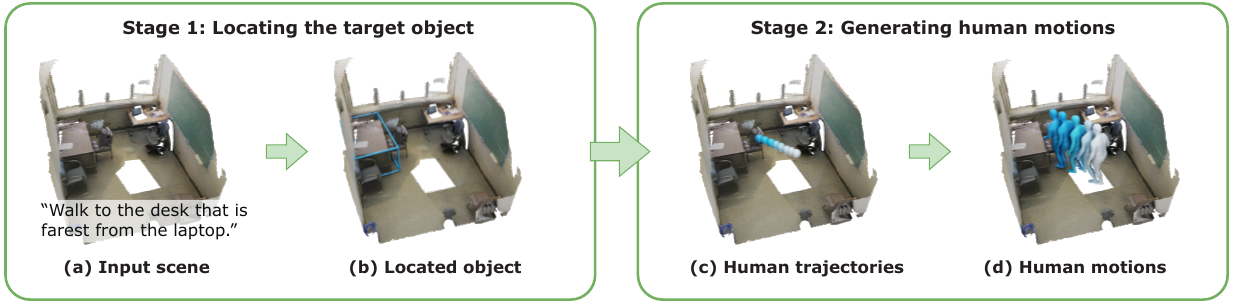}
    
    \caption{\protect\textbf{Overview of our two-stage pipeline.} 
    In the first stage, given an input scene and a text description (a), we use \protect\chatgpt{} to locate the target object (b). 
    In the second stage, human motions are synthesized by first producing human trajectories (c) and then generating local poses (d).
    }
    \label{fig:overview}
\end{figure*}
\input{sec/2_relatedwork}
\begin{figure*}
    \centering
    \includegraphics[width=0.85\linewidth]{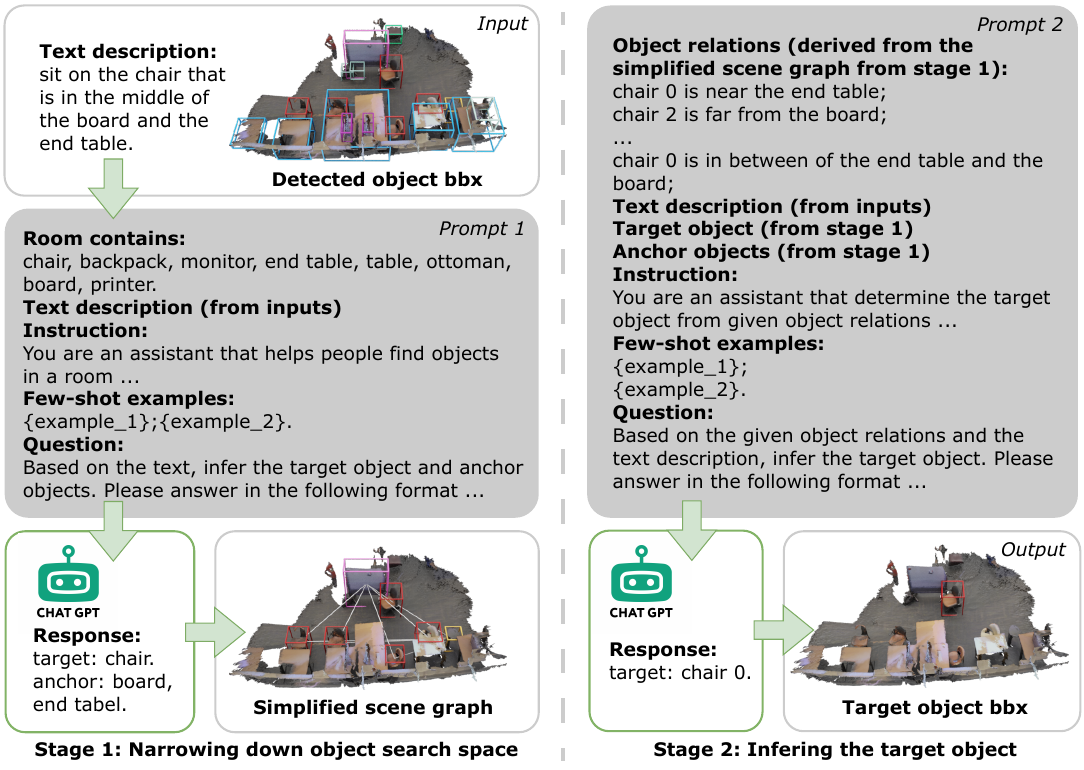}
    
    \caption{\protect\textbf{Pipeline of localizing the target object.}
    In stage 1, given the input text description and detected object bounding boxes (bbx), we construct the first prompt asking \protect\chatgpt{} the categories of target objects and anchor objects. Based on the response, the scene graph can be simplified. In stage 2, we construct the second prompt with inputs and results from stage 1, including object relations derived from the simplified scene graph. The second prompt is designed for asking \protect\chatgpt{} to infer the target object. Finally, we can get the target object bounding box from the response of \protect\chatgpt{}.
    }
    \label{fig:chatgpt}
\end{figure*}
\input{sec/3_preliminary}
\input{sec/4_method}
\input{tables/00_maintable}
\begin{figure*}[tbp]
    \centering
    \includegraphics[width=\linewidth]{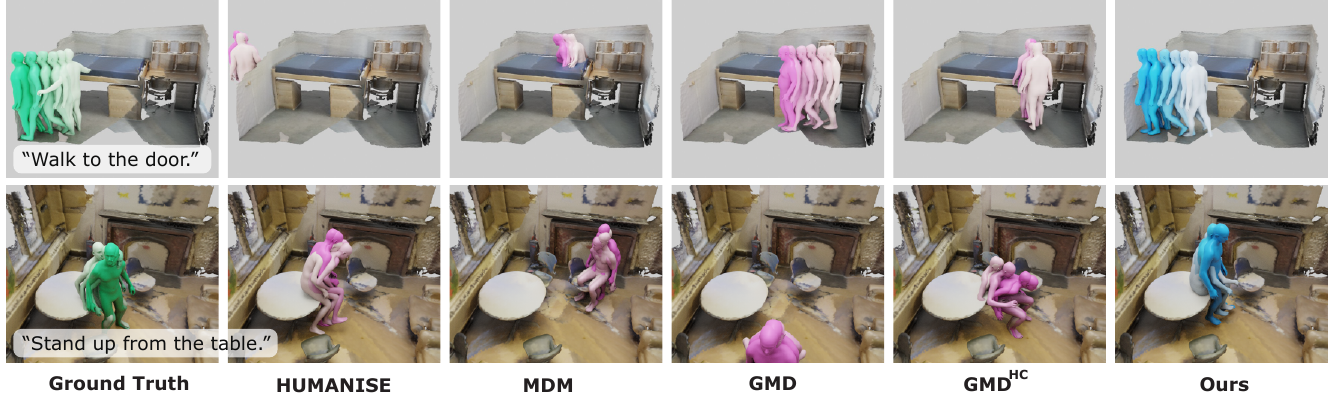}    

    \caption{\protect\textbf{Qualitative results.} 
    We compare our method with groundtruth and four baselines (please refer to \protect\secref{sec:compare}) given the same text descriptions. 
    Our method synthesizes motions that interact with the object precisely as the groundtruth data while the baselines fail.}
    
    \label{fig:quality}
\end{figure*}
\input{sec/5_experiments}
\input{tables/01_stage}
\input{tables/02_object}
\begin{figure}[t]
    \centering
    \includegraphics[width=0.90\linewidth]{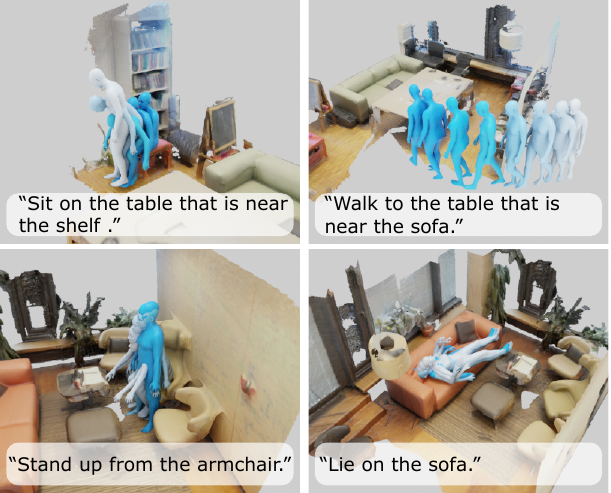}

    \caption{\protect\textbf{Qualitative results of our method on the PROX dataset.}
    We run our method on the scenes from the PROX dataset without fine-tuning.
    Results show that our method is capable to generalize to unseen scenes and objects.
    }
    \label{fig:prox}
\end{figure}
\input{sec/6_discussion}
\input{sec/7_conclusion}

\vspace{6pt}\noindent{\large\bf Acknowledgement}
\vspace{6pt} \\
\noindent This work was partially supported by the NSFC (No. 62172364), Ant Group and Information Technology Center and State Key Lab of CAD\&CG, Zhejiang University.

{
    \small
    \bibliographystyle{ieeenat_fullname}
    \bibliography{main}
}


\end{document}

%% file: preamble.tex
\usepackage[dvipsnames]{xcolor}

\usepackage{graphicx}
\usepackage{amsmath}
\usepackage{booktabs}
\usepackage{multirow}
\usepackage{colortbl}
\usepackage{makecell}
\usepackage{diagbox}

\newcommand{\figref}[1]{Fig.~\ref{#1}}
\newcommand{\tabref}[1]{Tab.~\ref{#1}}
\newcommand{\secref}[1]{Sec.~\ref{#1}}

\newcommand{\equref}[1]{Eq.~\ref{#1}}
\newcommand{\PAR}[1]{\vskip4pt \noindent{\bf #1~}}

\newcommand{\chatgpt}{ChatGPT}
\newcommand{\smplx}{SMPL-X}

\newcommand{\humanise}{HUMANISE}
\newcommand{\llama}{LLaMA~2}
\newcommand{\mdm}{$\text{MDM}^{*}$}
\newcommand{\gmd}{$\text{GMD}^{*}$}
\newcommand{\ghc}{$\text{GMD}^{\text{HC}}$}
\newcommand{\wolocal}{``w/o localization''}
\newcommand{\woobject}{``w/o object-centric''}
\newcommand{\woseparate}{``w/o two-stage''}
\newcommand{\wodiffusion}{``w/o diffusion''}
\newcommand{\wopretrain}{``w/o pretrain''}
\newcommand{\usematching}{``ours using text matching''}
\newcommand{\usellama}{``ours using LLaMA''}
\newcommand{\usemistral}{``ours using Mistral''}
\newcommand{\wotwostage}{``ours w/o two-stage''}
\newcommand{\wofewshot}{``ours w/o few-shot''}

\newcommand{\mulm}{multimodality}

%% file: figures/figure_teaser.tex
\begin{center}
    \includegraphics[width=\linewidth]{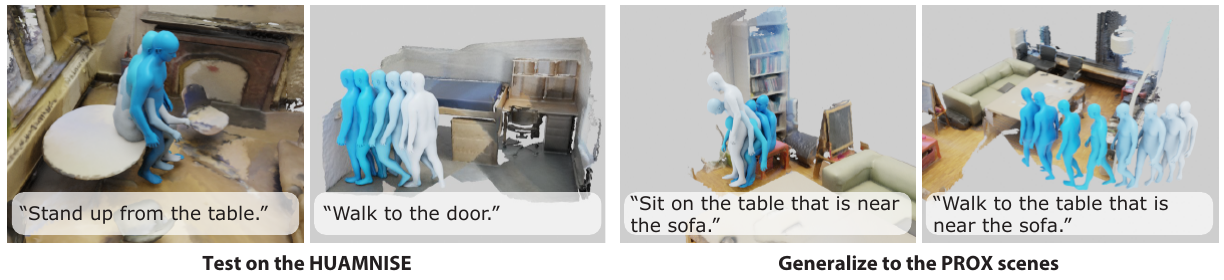}\vspace{-0.4cm}
\end{center}
\captionof{figure}{
    \textbf{Generating human motions in 3D scenes from text descriptions.} Our method can generate human motions containing accurate human-object interactions in 3D scenes based on textural descriptions. Although our method is trained and tested on the HUMANISE dataset, it can generalize to other scenes, e.g., the scenes in the PROX dataset. Left: test results on the HUMANISE dataset. Right: generalization results on the PROX scenes.}

%% file: sec/0_abstract.tex
\begin{abstract}
Generating human motions from textual descriptions has gained growing research interest due to its wide range of applications. 
However, only a few works consider human-scene interactions together with text conditions, which is crucial for visual and physical realism.
This paper focuses on the task of generating human motions in 3D indoor scenes given text descriptions of the human-scene interactions. 
This task presents challenges due to the multi-modality nature of text, scene, and motion, as well as the need for spatial reasoning. 
To address these challenges, we propose a new approach
that decomposes the complex problem into two more manageable sub-problems:
(1) language grounding of the target object and (2) object-centric motion generation. 
For language grounding of the target object, we leverage the power of large language models. 
For motion generation, we design an object-centric scene representation for the generative model to focus on the target object, thereby reducing the scene complexity and facilitating the modeling of the relationship between human motions and the object.
Experiments demonstrate the better motion quality of our approach compared to baselines and validate our design choices.
Code will be available at \href{https://zju3dv.github.io/text_scene_motion}{link}.
\end{abstract}

%% file: sec/1_intro.tex
\section{Introduction}
Human motion generation has been a long-standing problem due to its broad range of applications such as game development, virtual reality, and movie production. 
Recently, this area witnessed a paradigm shift from avatar animation given rich user input \cite{17tog_pfnn} to learning-based motion generation from high-level language prompts, e.g. text descriptions about the desired motion \cite{193dv_l2p, 21iccv_scatd, 22eccv_temos, 22eccv_motionclip, 22cvpr_humanml3d, 223dv_teach, 22eccv_tm2t, 22iclr_hmd, 23aaai_flame}.
However, most prior works on text-driven motion synthesis do not consider human-scene interactions \cite{22eccv_temos, 22eccv_motionclip, 22iclr_hmd, 23aaai_flame} while the scene context and physical constraints of the environment largely define the fidelity of the generated human motions.

In this paper, we focus on generating motions from text descriptions in 3D indoor scenes.
Specifically, given a 3D scan of the target scene and a text description of a human action that interacts with the scene, we aim to generate natural human motions that are consistent with the text description.

This problem presents several challenges, primarily due to the multi-modality nature of text, scene, and human motion.
In contrast to previous methods \cite{193dv_l2p, 22eccv_temos, 22eccv_motionclip, 223dv_teach,22iclr_hmd} 
focusing solely on textual descriptions of how human moves, our task also includes texts that additionally describe the spatial details in the given scene (e.g., sit on the armchair near the desk).
Therefore, this task requires spatial reasoning skills \cite{20eccv_referit3d}, where the model should build a text-object mapping to locate a specific object in 3D scenes aligned with a natural language description. 
In addition, the generated motions should also be coherent with scene contexts.

As a pioneer work, \humanise{} \cite{22nips_humanise} builds a conditional variational autoencoder (cVAE) \cite{14iclr_vae, 15nips_cvae} with separate encoders of scenes and texts for multi-modality understanding.
To enable spatial reasoning ability, they introduce auxiliary tasks of directly regressing object centers to learn 3D visual grounding in an implicit manner.
But they do not explicitly utilize the predicted centers and thus the inductive bias of visual grounding cannot be fully incorporated.
In addition, \humanise{} \cite{22nips_humanise} encodes the entire scene with a single point transformer \cite{21iccv_pointtransformer}.
Directly generating motions with such a model is challenging, as 3D point clouds are inherently noisy and complex \cite{21iccv_sat, 22cvpr_3dsps}, leading to the inability to locate the target object.
As suggested by \cite{23arxiv_continualmotion}, not every point of the scene is relevant to the final human motions.
Therefore, it is also necessary to develop a more targeted approach that focuses on the relevant parts of the scene to improve the quality of motion generation.

To tackle the problems mentioned above, we propose a novel approach that exploits the power of the large language models (LLMs) \cite{20nips_llmfewshot, 22nips_llmzeroshot, 22arxiv_chatgpt}.
Our key idea is to address the challenging task of generating motion in a scene based on textual cues by breaking it down into two smaller problems:
(1) language grounding of objects in 3D scenes and (2) generating motions with a focus on the target object.
For language grounding, rather than directly learning text-object mapping, we propose to formulate it as question answering and utilize the large prior knowledge of LLMs.
Specifically, we first construct scene graphs of 3D scenes and generate their textual descriptions.
Then we employ \chatgpt{} \cite{22arxiv_chatgpt} to analyze the relationship between scene descriptions and input instructions and respond with 3D visual grounding answers.
Experiments prove the effectiveness of this strategy.

For motion generation, we design an object-centric representation to help the generative model focus on the target object.
Specifically, we convert point clouds around the target object into volumetric sensors \cite{19tog_nsm} to build object-centric representation.
Then we employ diffusion models \cite{22iclr_hmd, 23iccv_hghoi} to synthesize human motions given object-centric representation and texts.
Compared with original scene point clouds which might have various scales, object-centric representation is more compact and robust to scales as objects in the same category are of similar size.
Therefore, this representation reduces scene complexity and facilitates the modeling of the relationship between motions and objects.
We conducted thorough comparative and ablation experiments on the \humanise{} dataset.
The results demonstrate that our method outperforms the baseline, reflected in more accurate object grounding results and better motions that align with the textual descriptions and scenes.
We further show that our approach can generalize to the PROX dataset \cite{19iccv_prox} without any fine-tuning.

%% file: sec/2_relatedwork.tex
\section{Related work}
\subsection{Motion synthesis}
\PAR{Deep learning for motion synthesis.} 
Deep learning methods for motion synthesis have attracted increasing attention in recent years \cite{15iccv_erd, 16tog_dlfcmse, 17cvpr_resrnn, 17tog_pfnn, 18tog_mann, 20tog_localphase, 20tog_rmi, 22nips_nemf}.
Various techniques, including MLP \cite{17tog_pfnn}, mixture of experts (MoE) \cite{18tog_mann, 22tog_deepphase}, and recurrent neural network (RNN) \cite{20tog_rmi}, are employed to tackle this task.
To generate diverse motion results, previous works explore cVAE \cite{20tog_motionvae, 22cvpr_gamma}, GANs \cite{22tog_ganimator, 23cvpr_modi}, normalizing flow \cite{20tog_moglow}, and diffusion models \cite{22iclr_hmd}.
%
%
\PAR{Text-driven motion generation.}
Recently, there has been a growing interest in text-driven motion generation \cite{193dv_l2p, 22eccv_temos, 22eccv_motionclip}.
This task takes natural language as input and synthesizes human motions that align with instructions.
KIT-ML \cite{16bigdata_kit} is the first benchmark of this task.
Some works further annotate the AMASS dataset \cite{19iccv_amass} with action labels \cite{21cvpr_babel} and text \cite{22cvpr_humanml3d}.
To tackle this task, \cite{193dv_l2p, 21iccv_scatd, 22eccv_motionclip} propose to learn a shared latent space for text and motion.
\cite{22eccv_temos} employs transformer VAE \cite{21iccv_actor} to generate diverse results.
\cite{22eccv_tm2t, 23cvpr_discreterepr} achieve better performance by discrete representation with VQ-VAE \cite{17nips_vqvae, 19nips_vqvae2}.
\cite{22iclr_hmd, 23aaai_flame, 22arxiv_motiondiffuse, 23cvpr_mofusion} successfully apply diffusion models in this direction.
\cite{23cvpr_belfusion, 23iccv_priority} further explore latent diffusion models \cite{22cvpr_stablediffusion}.
Based on \cite{22iclr_hmd}, \cite{23iccv_gmd, 23arxiv_omnicontrol} introduce sparse spatial control.
\cite{223dv_teach, 23cvpr_elaborative, 23arxiv_hmdasgp} also investigate long sequence motion generation.
Other existing works \cite{223dv_rachoi, 23icme_actiongpt, 23iccv_sinc, 23arxiv_motiongpt} leverage Large Language Models (LLM) \cite{23arxiv_llama, 23arxiv_gpt4, 20nips_llmfewshot, 22nips_llmzeroshot} in the motion domain.
\cite{23iccv_sinc} construct compositional motions by combining different body parts with LLM.
\cite{23arxiv_motiongpt, 23arxiv_motiongpt_tencent} regard motions as a kind of language and finetune LLMs \cite{20jmlr_t5} with motion tokens \cite{22eccv_tm2t, 23cvpr_discreterepr}.
%
%
\PAR{Scene-aware motion generation}
This direction is to generate human motions in a 3D scene \cite{19tog_nsm, 21cvpr_slt}.
To synthesize walking and sitting motions, \cite{19tog_nsm} construct volumetric sensors to encode object information and the surroundings of the character.
\cite{21iccv_samp} extends it to synthesize motions with diverse sitting and lying styles.
Furthermore, \cite{22eccv_couch} proposes to control sitting styles by hand contacts.
To improve the performance, \cite{23iccv_hghoi} employs a hierarchical framework that generates goal poses, milestones, and motion sequentially.
\cite{22eccv_saga, 22cvpr_goal} synthesize whole-body (body and hand fingers) grasping motions with small objects.
\cite{20tog_localphase, 23cgf_imos, 23arxiv_toho, 23iccv_interdiff} explore interactions with dynamic objects including manipulation \cite{23cgf_imos, 23arxiv_toho} and carrying \cite{23iccv_interdiff}.
\cite{21aaai_learning2sit, 23arxiv_phycsi} explore human-object interactions with physically simulated characters.
These works mainly focus on one or two objects while others \cite{21cvpr_slt, 22cvpr_tdns} consider a more complex scene input (e.g., point clouds of the scene including walls and floors.).
\cite{21cvpr_slt, 22cvpr_tdns} adopt hierarchical frameworks and generate trajectory and motion separately.
\cite{22eccv_gimo} introduces gaze to help generation.
\cite{23cvpr_scenediffuser} further employs diffusion models and \cite{23arxiv_continualmotion} synthesizes very long-term motions in scenes by dividing long sequences into several short sequences.
\cite{23iccv_dimos} designs a reinforcement learning pipeline to enable navigation in a complex scene and interaction style control.
%
%
\PAR{Text-driven scene-aware motion generation}
Only a few works consider text and scenes simultaneously \cite{22nips_humanise, 23iccv_dimos, 23iccv_gmd}. 
Although \cite{23iccv_dimos} enables text control of sitting styles by \cite{2022cvpr_coins}, the text descriptions in our setting are used to select an object in a cluttered scene.
\cite{23iccv_gmd} could avoid obstacles during walking while our task needs to handle various actions.
The most relevant work to us is \humanise{} \cite{22nips_humanise}, which employs a transformer VAE architecture with a two-stream condition module for text and scenes.
To accurately localize target objects, they design auxiliary tasks like directly regressing object centers.
In contrast to \humanise, we propose a two-stage pipeline where we first localize the target object with the help of \chatgpt{} \cite{22arxiv_chatgpt} and then generate human motion using the object-centric representation.

\subsection{3D visual grounding.}
In recent years, visual grounding in 3D scenes has been explored \cite{20eccv_referit3d, 21corl_lg3d, 21iccv_yourefit} and also tackled in 3D question answering \cite{18corl_eqa, 22tvcg_3dqa}.
Given the point cloud of 3D scenes, this task \cite{20eccv_referit3d} requires models to locate the target object according to text instructions.
Most works follow a two-stage scheme where multiple bounding boxes \cite{20eccv_scanrefer, 22corl_languagerefer, 21iccv_3dvgtransformer} or segmentation results \cite{21aaai_ggnnref3Dinst, 21iccv_instancerefer} are first predicted and then selecting the object according to language descriptions.
\cite{22cvpr_mvtrans3dgrounding} introduces multi-view inputs, and \cite{21iccv_sat} employs 2D semantics.
\cite{22eccv_butddetr} combines bottom-up \cite{21iccv_groupfree3ddet} and top-down \cite{20eccv_detr} detection methods.
\cite{22cvpr_3dsps} designs a single-stage pipeline by progressively selecting key points.
\cite{23iccv_viewrefer} extends \cite{22cvpr_mvtrans3dgrounding} with the help of GPT \cite{20nips_llmfewshot} to generate multi-view text inputs.
More recently, \cite{23cvpr_ns3d} proposes a neuro-symbolic framework with large language-to-code models \cite{21arxiv_codex}.
Most works only localize a single object and \cite{23iccv_multi3drefer} could localize a flexible number of objects.
Different from previous works which directly handle point clouds or multi-view images, we convert the scene into textual descriptions and leverage large language models to infer the target object.
Like \cite{23cvpr_ns3d}, in this work, we also leverage large language models for object localization.

%% file: sec/3_preliminary.tex
\section{Problem setup and preliminaries}
\label{sec:setup}
In this section, we discuss the definition of the task and preliminaries.
We aim to generate human motion that is consistent with both the text description and the given scene. 

\PAR{Text descriptions.}
The text description follows the template in Sr3D \cite{20eccv_referit3d}
(e.g., ``$<$sit on$>$ $<$the chair$>$ [$<$in the center of$>$ $<$desk and bookshelf$>$]'').
There are four actions (walk, sit, stand up, and lie).
The $target$ represents the object that the agents need to interact with,
and the $anchor$ objects help to determine the $target$.
A certain type of target furniture category usually has many instances in one scene, while the anchor furniture should be unique.
To specify the exact target object, there are five types of spatial relations \cite{20eccv_referit3d} between the $target$ and the $anchor$: horizontal proximity, vertical proximity, between, allocentric, and support. 

\PAR{Scene representations.} The scene is denoted as a point cloud of $N$ points: $ S \in \mathbf{R}^{N \times 6}$, containing the position and normal direction information of each point. 

\PAR{Motion representations.} The output motion sequence is represented as a sequence of \smplx{} \cite{19cvpr_smplx} body meshes $M$. \smplx{} is a parametric human body model. In this work, body parameters include body shape parameters $\beta \in \mathbf{R}^{10}$, global translation $r \in \mathbf{R}^3$, 6D global orientation $ \gamma \in \mathbf{R}^6$, and 6D pose parameters of $J$ joints $\boldsymbol{\theta} \in \mathbf{R}^{J\times6}$ \cite{19cvpr_6drot}. Following \cite{22nips_humanise}, $\beta$ is treated as a condition to model the effect of body shape, and we omit it for ease of notation. Note that start position and pose are also generated by the model.

\PAR{Diffusion models.}
The diffusion \cite{20nips_ddpm} is defined as a Markov noising process $\left\{\mathbf{x}_t\right\}_{t=0}^T$ that follows $q\left(\mathbf{x}_t \mid \mathbf{x}_0\right)$, where $\mathbf{x}_0 \sim q\left(\mathbf{x}_0\right)$ is the data and $\mathbf{x}_t$ is the noised data at noising step $t$.
The formal definition is:
\begin{equation}
    q\left(\mathbf{x}_t \mid \mathbf{x}_0\right)=\mathcal{N}\left(\mathbf{x}_t ; \sqrt{\bar{\alpha}_t} \mathbf{x}_0,\left(1-\bar{\alpha}_t\right) \mathbf{I}\right),
\end{equation}
where $\bar{\alpha}_t$ are constants with monotonically decreasing schedule.
When $\bar{\alpha}_t$ is small enough, we can approximate $\mathbf{x}_T \sim \mathcal{N}(\mathbf{0}, \mathbf{I})$.
In our context, we use conditional diffusion models like \cite{22iclr_hmd, 22arxiv_dalle2}. The training loss is defined as:
\begin{equation}
\label{eq:simple}
    \mathcal{L}=\mathrm{E}_{t \in[1, T], \mathbf{x}_0 \sim q\left(\mathbf{x}_0\right)}\left[\|\mathbf{x}_0 - G(\mathbf{x}_t, t, \mathbf{C})\|\right],
\end{equation}
where $G$ is the generative model, and $\mathbf{C}$ is the condition.

%% file: sec/4_method.tex
\section{Method}
\input{figures/figure_sensor}
The overview of our method is shown in \figref{fig:overview}.
In \secref{sec:object}, we leverage \chatgpt{} to localize the target object given a 3D scene and a textual description.
Based on the accurate localization, we can focus on the target object and employ an object-centric generation pipeline to separately synthesize trajectories (\secref{sec:traj}) and motions (\secref{sec:motion}).
Implementation details are discussed in \secref{sec:implementation}.

\subsection{Language grounding of objects in 3D scenes}\label{sec:object}
Scene-aware motion generation from textual descriptions requires the scene-understanding ability and build the relationship between scenes and texts.
Since the instructions describe how the character moves and interacts with a single target object, the majority of the scene might bear negligible relevance to the final motions \cite{23arxiv_continualmotion}.
Motivated by this, we propose to locate the target object to identify the most pertinent information.
Inspired by the recent progress of LLM \cite{22nips_llmzeroshot, 23arxiv_gpt4, 22arxiv_chatgpt}, \chatgpt{} \cite{22arxiv_chatgpt} is utilized to find the specific objects in the given text.
We first obtain textual descriptions of scenes by building scene graphs.
Then we feed them with text instructions to \chatgpt{} with specially designed prompts and parse the response to get target objects.

\PAR{Spatial scene graph extraction.} 
To utilize LLM, the initial step involves converting a 3D scene into text.
This is achieved by building a \textit{spatial scene graph}.
We utilize a pre-trained 3D object detection model in \cite{21iccv_groupfree3ddet} to provide 3D box proposals.
Subsequently, we follow the approach of \cite{20eccv_referit3d} to obtain the relationships between objects.
Specifically, for every set of two objects, we can infer their relationships in three types from their bounding boxes: \textit{horizontal proximity}, \textit{vertical proximity}, and \textit{support} as mentioned in \secref{sec:setup}; for every set of three objects, we can infer if one of them is in \textit{between} of the other two objects.
As detection results do not contain object poses, we do not construct \textit{allocentric} relations (e.g., ``the shelf that is behind the sofa'').
Then, we build a scene graph, where each object is assigned as a node in the graph, and edges between nodes represent the relationships between objects.
By converting 3D scenes into text, we can apply \chatgpt{} to extract meaningful insights from the data.

\PAR{Leveraging \chatgpt{} to localize the target object.}
A simple approach is to directly input the entire scene graph into \chatgpt{} and ask it to select the target objects.
However, scenes may contain many objects, resulting in extremely long textural descriptions.
We observe that \chatgpt{} is often confused in such settings and fails to respond with the right answer.
To narrow the search space, we first employ \chatgpt{} to recognize objects that correlate to the provided text.
We then exclude the unrelated objects from the scene graph, focusing solely on those with pertinent information, enabling us to pinpoint the target object effectively.
This approach has the advantage of reducing the number of objects that need to be considered, making it easier for \chatgpt{} to identify the target object. 

As shown in \figref{fig:chatgpt}, we construct two prompts in sequence. 
Take ``sit on the chair that is in the middle of the board and the end table'' as an example, we first need to narrow down the object category search space. 
In order to find which type of objects we care about, we construct the first prompt that asks \chatgpt{} to find out \textit{target objects} and \textit{anchor objects}.
\textit{Target object} is defined as the final object that we want the agent to interact with, which in this case, is the ``table''. \textit{Anchor objects} are the objects that help with determining the \textit{target object}, for there might be many chairs in one scene.
Based on the target object ``chair'' and anchor objects ``board'' and ``end table'', we can filter out all the unrelated objects in the scene, only keeping chairs, the board, and the end table in our scene graph.
Next, according to the simplified scene graph, we can describe object relations in text: every edge in the scene graph could be converted to an \textit{edge sentence} like ``chair 4 is far from the end table 0''. Converting all edges to such edge sentences gives a full description of the current scene.
Finally, we construct a second prompt by asking \chatgpt{} to infer the target object from the accumulated edge sentences.

\subsection{Diffusion-based trajectory generation} \label{sec:traj}
Given the localized object from \chatgpt, we first generate the trajectory based on the instructions and then synthesize local human poses.
Trajectory is defined as a sequence of characters' translations and orientations.
As suggested by \cite{23arxiv_continualmotion}, not every point of the scene is relevant to the final human motions.
Inspired by NSM \cite{19tog_nsm} and ManipNet \cite{21tog_manipnet}, we employ volumetric sensors (as shown in \figref{fig:sensor}) around the target object to represent the scene.

\PAR{Object-centric scene representation.}
Denote the target object center location as $c_o = (c_x, c_y, c_z)$.
We transform the point cloud of the scene $S$ to an object-centered coordinate axis centered at $c_o$.
To recognize the surrounding geometry of the target object, we create volumetric sensors called Environment Sensor and Target Sensor.

\PAR{Environment sensor.}
The environment sensor is centered at $c_o$ in a cubic shape with a volume of $4\times 4\times 4$ $m^3$, containing $8\times 8\times 8$ cubic voxels as shown in \figref{fig:sensor} (c).
It is constructed by collecting all occupancies $o_s$, center positions $c_{v}$, and normal directions $n_{v}$ of each voxel to form a feature vector $E$.
Like \cite{19tog_nsm}, the scene occupancy $o_s \in R^1$ in each voxel is defined based on the given scene mesh:
\begin{equation}
\label{equ:environment_sensor}
    o_s = 
    \begin{cases}
        1 & \text{if } d_s < 0,\\
        0 & \text{if } d_s > a_s, \\
        1 - \frac{d_s}{a_s} & \text{otherwise},
    \end{cases}
\end{equation}
where $d_s$ is the signed distance between the scene mesh and the voxel center, and $a_s$ is the voxel edge length.
$n_{v}$ is the normal direction of the closest scene point to the voxel center.
The environment sensor provides coarse scene geometry around the target object.

\PAR{Target sensor.}
To capture the detailed geometry of the target object, we further build a target sensor.
As we already obtain the 3D bounding box of the target object in \secref{sec:object}, we crop the point clouds according to the bounding box and construct an $8 \times 8 \times 8$ cubic volumes that cover the bounding box, as shown in \figref{fig:sensor} (b).
Target sensor $T$ is in the same form as the environment sensor $E$, except the target sensor has a different voxel size, as visualized in \figref{fig:sensor} (b) and (c).

Given the constructed object-centric representations, we follow \cite{23iccv_hghoi} to employ a transformer decoder architecture \cite{17nips_attention}, which enables arbitrary length motions.
As for the text input, we use CLIP \cite{21icml_clip} text encoder to encode input text to the text feature $L$.
The time-step $t$ is injected into the decoder in sinusoidal position embeddings form \cite{17nips_attention}.
In summary, the condition for this generation model is 
\begin{equation}
\label{eq:traj_c}
    \mathbf{C_t} = \{L, E, T\},
\end{equation}
where $L$ is the text feature, $E$ is the environment sensor, and $T$ is the target sensor.
All the conditions are projected to the same dimension $D = 512$ by MLPs and summed with positional embeddings to form tokens.
We use the simple objective described in \equref{eq:simple} to train the trajectory generation model $G_r$ to generate the trajectory $\mathbf{r}_{1:N}$ with length $N$.

\subsection{Diffusion-based motion completion}\label{sec:motion}

Given the trajectory from \secref{sec:traj}, the next step is to complete the whole motion. 
Based on the generated trajectories, we construct a Trajectory Sensor $O$ to explicitly reason about the interaction between the character and scenes.

\PAR{Trajectory sensor.}
The trajectory sensor is also a volumetric sensor which is similar to the form of the environment sensor but is designed for ego-centric perception \cite{22eccv_couch} as shown in \figref{fig:sensor} (d).
Specifically, trajectory sensors are positioned around characters in each frame.
This sensor $O_i$ is centered at the predicted root position of the $i$-th frame and faces to the $i$-th frame's predicted root orientation, containing $8\times 8\times 8$ cubic voxels that store scene occupancy.
The occupancy calculation is the same as \equref{equ:environment_sensor}.

Another transformer-based conditional diffusion model is used to synthesize local poses along trajectories.
The condition is defined as:
\begin{equation}
    \mathbf{C_m} = \{L, E, T, O_1, ..., O_N\},
\end{equation}
where $L$, $E$, and $T$ have the same meanings in \equref{eq:traj_c}.
Simple objective described in \equref{eq:simple} is used to train the motion generation model $G_m$ to generate global orientation $\gamma_{1:N}$ and local poses $\boldsymbol{\theta}_{1:N}$ with length $N$.

\subsection{Implementation details}\label{sec:implementation}
Following \cite{21cvpr_slt, 22cvpr_tdns, 23iccv_hghoi, 23iccv_gmd}, we use separate diffusion models for the trajectory generation and the motion generation.
Because the HUMANISE dataset \cite{22nips_humanise} contains a relatively small number (51 minutes) of pure motion samples from the AMASS dataset (3772 minutes) \cite{19iccv_amass}, we first pre-train the models on the whole AMASS dataset for $200$ epochs and then fine-tune on the HUMANISE dataset for $200$ epochs to improve the motion quality.
During pretraining, the text feature $L$ is set to all zeros.
Both models are trained with the AdamW optimizer \cite{14arxiv_adam}, using a learning rate of $0.0001$ on a single Nvidia RTX 3090 GPU.
The batch size is set to 128.
The version of \chatgpt{} is \textit{gpt-3.5-turbo}.
More details can be found in the supplementary material.

%% file: figures/figure_sensor.tex
\begin{figure}[tbp]
    \centering
    \includegraphics[width=0.95\linewidth]{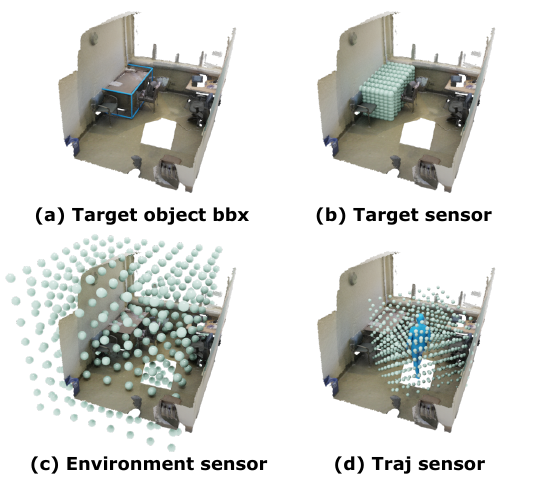}
    
    \caption{\protect\textbf{The visualization of the environment sensor, target sensor, and trajectory sensor.}
    The target sensor (b) gives detailed geometry of the target object. The environment sensor (c) gives coarse spatial information around the target object. The trajectory sensor (d) is located around the human.}
    \label{fig:sensor}
\end{figure}

%% file: tables/00_maintable.tex
\begin{table*}[tbp]
\centering
\resizebox{0.85\linewidth}{!}
{
\begin{tabular}{ccccccccc}
    \toprule
    \multirow{2}{*} & \multirow{2}{*}{Methods} & \multicolumn{2}{c}{Scene-conditional} & \multicolumn{3}{c}{Action-conditional} & \multicolumn{2}{c}{Pure motion quality} \\    
    \cmidrule(r){3-4} \cmidrule(r){5-7} \cmidrule{8-9}    
    
    & & goal dist.$\downarrow$ & scene score$\uparrow$ & accuracy$\uparrow$ & diversity$\rightarrow$ & \mulm$\rightarrow$ & FID$\downarrow$ & quality score$\uparrow$ \\
    
    \midrule
    \multirow{6}{*}
    & Real        & 0.014 & - & 99.1\% & 4.82 & 2.28 & 0.00 & - \\
    & \mdm        & 2.048 & 1.99 & 96.0\% & \textbf{4.83} & 2.57 & \underline{0.16} & 2.45 \\
    & \gmd        & 1.229 & \underline{2.50} & \underline{96.6\%} & 4.91 & \textbf{2.28} & 0.24 & 2.64 \\
    & \ghc        & 1.130 & 2.49 & 96.4\% & 4.93 & 2.43 & 0.23 & 2.76 \\
    & \humanise{} & \underline{0.995} & 2.33 & 89.7\% & 4.25 & 2.66 & 1.11 & \underline{2.81} \\
    \midrule
    & Ours        & \textbf{0.384} & \textbf{3.54} & \textbf{97.1\%} & \underline{4.78} & \underline{2.31} & \textbf{0.12} & \textbf{3.50} \\
    
    \bottomrule
\end{tabular}
}

\caption{\protect\textbf{Quantitative results on the HUMANISE dataset.} We compare our method with four baselines (please refer to \protect\secref{sec:compare}) and the real data. \protect\ghc{} means using \protect\humanise{}'s predicted center to guide motion generation in \protect\gmd. Among the metrics, scene score and quality score are perceptual studies. $\uparrow$ means higher is better and $\downarrow$ means lower is better. $\rightarrow$ means closer to the real data is better. \protect\textbf{Bold} indicates the best results. \protect\underline{Underline} indicates the second best.
}
\label{tab:maintable}
\end{table*}

%% file: sec/5_experiments.tex
\section{Experiments}

\subsection{Evaluation metrics}\label{sec:metrics}
We evaluate the generated motions in three aspects: scene-conditional, action-conditional, and pure motion quality.

\PAR{Scene-conditional motion quality.}
To evaluate how the generated motion is aligned with the scene, we calculate the body-to-goal distance (\textbf{goal dist.})~\cite{22nips_humanise} to measure how accurately the character interacts with target objects.
However, goal dist.~does not consider the consistency of the entire motion and scene (e.g., sitting on the sofa with incorrect orientation).
To compensate for goal dist., a human perceptual study (indicated by \textbf{scene score}) is performed by randomly sampling 20 scenarios for each model, where ten workers are required to score each sample.

\PAR{Action-conditional motion quality.}
To measure how the generated motion is aligned with the text, we follow \cite{20mm_action2motion} to evaluate action recognition accuracy (\textbf{accuracy}), \textbf{diversity}, and \textbf{\mulm} of the results.
Calculating these metrics relies on a pre-trained action recognition model \cite{18aaai_stgcn} and we train the recognition model on the \humanise{} dataset.

\PAR{Pure motion quality.}
We evaluate the realism of generated motions using the metric suggested by \cite{20mm_action2motion}, namely Frechet Inception Distance (\textbf{FID}).
A lower FID indicates that generated motions are closer to the groundtruth motions.
We also perform a human perceptual study (indicated by \textbf{quality score}) to measure pure motion quality.

\subsection{Generating human motion from text and scene} \label{sec:compare}
Our experiments are conducted on the \humanise{} dataset \cite{22nips_humanise}.
Following the previous setting \cite{22nips_humanise}, there are $16.5k$ motion sequences in $543$ scenes for training and $3.1k$ motion sequences in $100$ scenes for testing.
We compare our method with four baselines: 
(1) \textbf{\mdm}~\cite{22iclr_hmd}: a diffusion-based motion generator. 
(2) \textbf{\gmd}~\cite{23iccv_gmd}: GMD proposes various techniques for enhancing the control and quality of MDM. We employ the single-stage setting of \gmd. 
Similar to \humanise, we use a point-transformer to provide scene features for \mdm and \gmd.
(3) \textbf{\ghc}: we provide object centers predicted by \humanise{} (denoted by $\text{HC}$) to guide the motion generation process in GMD by adding a proximity loss to encourage the motion to be close to the predicted object center. The proximity loss is defined as the distance from $\text{HC}$ to the predicted human pelvis at the interacting frame.
(4) \textbf{\humanise}: we directly use their released models.

The quantitative results are shown in \tabref{tab:maintable}.
Our method outperforms the baseline in terms of goal dist., scene score, accuracy, FID, and quality score and achieves competitive results in diversity and \mulm.
The quantitative results show that our approach can generate better-quality motions and are more consistent with all the conditions. 
The qualitative results are demonstrated in \figref{fig:quality}.
Please refer to the supplementary material for more visualizations.

\subsection{Ablation study} \label{sec:ablation}
\PAR{Ablation of main components.}
Four variants are constructed to explore the effect of our design choice.
(1) \textbf{\wolocal}: the localization module is removed and trajectories and motions are generated directly using scene point clouds. 
Scene features are given as the same form of \mdm and \gmd.
(2) \textbf{\woobject}: we remove our object-centric representation and predict motions in the scene coordinate. The scene features of our sensors are still employed.
(3) \textbf{\woseparate}: trajectories and motions are generated together.
(4) \textbf{\wodiffusion}: we employ cVAE instead of the diffusion model as the architecture.
(5) \textbf{\wopretrain}: the models are not pre-trained on the AMASS dataset.
As shown in \tabref{tab:stage}, the localization enhances the precise interaction ability and the object-centric representation improves motion quality. 
Although the (\woseparate) is more efficient and has competitive performance metrics comparing to our method, it is prone to collapse directly into the target position, lacking gradual transition and realism.

\PAR{The design choice of object localization.}
For object localization, we compare our method with two baselines and five variants.
(1) \textbf{\humanise}: predicts the object center in the auxiliary task.
(2) \textbf{BUTD-DETR}~\cite{22eccv_butddetr}: is a 3D visual grounding method.
(3) \textbf{\wotwostage}: we employ a one-stage question-answering process.
(4) \textbf{\wofewshot}: the few-shot examples are not provided.
(5) \textbf{\usematching}: we compute CLIP similarities instead of using \chatgpt.
(6) \textbf{\usellama}: \chatgpt{} is replaced by \llama~7B model.
(7) \textbf{\usemistral}: \chatgpt{} is replaced by Mistral 7B \cite{23arxiv_mistral}, an open-source LLM.
The prompts for \llama and Mistral 7B are slightly adjusted.
We evaluate these variants under two scenarios: predicted detection \cite{21iccv_groupfree3ddet} and groundtruth detection.
The metrics include the accuracy (\textbf{acc.}) and the distance from predicted centers to the object center (\textbf{center dist.}). 
``acc.'' is the percentage of times with an IoU (Intersection over Union) higher than the threshold (0.25) following \cite{22eccv_butddetr}.
The results are shown in \tabref{tab:object}.
Since the results of Mistral are only slightly behind of ChatGPT,
ChatGPT can be replaced by Mistral for better reproducibility,
but cannot be replaced by the text matching method.
Moreover, if groundtruth detection is provided, our method can achieve even better results.

\PAR{The design choice of sensor density.}
Please refer to the supplementary material.

\subsection{Generalization}
To validate the generalization ability of our method, we run our method directly on the unseen PROX dataset \cite{19iccv_prox} without fine-tuning. We provide illustrations in \figref{fig:prox}. 
With our localization method based on \chatgpt{} and motion generation method based on volumetric sensors, our pipeline can easily generalize to other datasets.
Please refer to the supplementary material for more results.

%% file: tables/01_stage.tex
\begin{table}[tbp]
\centering
\resizebox{\linewidth}{!}
{
\begin{tabular}{cccccc}
    \toprule
    \multirow{2}{*}{Variants} & \multicolumn{1}{c}{Scene} & \multicolumn{3}{c}{Action} & \multicolumn{1}{c}{Pure} \\ 
    \cmidrule(r){2-2} \cmidrule(r){3-5} \cmidrule(r){6-6}
    & goal dist.$\downarrow$ & accuracy$\uparrow$ & diversity$\rightarrow$ & mm$\rightarrow$ & FID$\downarrow$ \\
    \midrule
    
    Real               & 0.014 & 99.1\% & 4.82 & 2.28 & 0.00 \\
    w/o localization   & 0.592 & 74.5\% & 3.71 & 2.75 & 3.14 \\
    w/o object-centric & 0.413 & 90.6\% & 4.13 & 2.63 & 1.01 \\
    w/o two-stage      & \underline{0.385} & \textbf{97.2\%} & \underline{4.93} & \underline{2.33} & \underline{0.14} \\
    w/o diffusion      & 0.390 & 40.0\% & 3.69 & 3.86 & 3.61 \\
    w/o pretrain       & 0.392 & 82.6\% & 3.88 & 2.37 & 2.50 \\
    \midrule
    Ours               & \textbf{0.384} & \underline{97.1\%} & \textbf{4.78} & \textbf{2.31} & \textbf{0.12} \\
    \bottomrule
\end{tabular}
}

\caption{\protect\textbf{Ablation of main components.} We compare our method with five variants (please refer to \protect\secref{sec:ablation}). 
Among them, \protect\textit{mm} indicates \protect\mulm. 
\protect\textbf{Bold} indicates the best results. \protect\underline{Underline} indicates the second best.
}
\label{tab:stage}
\end{table}

%% file: tables/02_object.tex
\begin{table}[tbp]
\centering
\resizebox{\linewidth}{!}
{
\begin{tabular}{ccccc}
    \toprule
    \multirow{2}{*}{Variants} & \multicolumn{2}{c}{Predicted detection} & \multicolumn{2}{c}{GT detection}\\
    \cmidrule(r){2-3} \cmidrule(r){4-5}

    & acc. $\uparrow$ & center dist.$\downarrow$ & acc. $\uparrow$ & center dist.$\downarrow$\\
    \midrule
    HUMANISE & - & 1.48 & - & -\\
    BUTD-DETR & 62.5 & 1.33 & 63.1 & 1.23\\
    \midrule
    Ours w/o two-stage & 49.0 & 1.34 & 58.5 & 1.10 \\
    Ours w/o few-shot & 69.3 & 0.80 & 86.3 & 0.45 \\
    Ours using text matching & 67.0 & 0.89 & 79.7 & 0.52 \\
    Ours using \llama & 40.3 & 1.55 & 42.5 & 1.52 \\
    Ours using Mistral 7B & 72.4 & 0.74 & 87.2 & \textbf{0.33} \\
    \midrule
    Ours using \chatgpt-3.5 & \textbf{75.6} & \textbf{0.61} & \textbf{90.5} & 0.37 \\
    \bottomrule
\end{tabular}
}

\caption{\protect\textbf{Design choices of object localization.} 
We compare our method with two baselines and five variants (please refer to \protect\secref{sec:ablation}). 
Since \protect\humanise{} directly regresses the coordinate of centers without utilizing groundtruth detection, we do not calculate acc. and only include their results under predicted detection. \protect\textbf{Bold} indicates the best results.
}
\label{tab:object}
\end{table}

%% file: sec/6_discussion.tex
\section{Discussion}
We have demonstrated that our approach could synthesize motions with better quality and more precise interactions.
However, restricted by the dataset, the duration of motions is relatively short (60\% are around 1-3s), and most texts follow the template \cite{20eccv_referit3d} without detailed descriptions.
We only tackle static scenes and extending our method to interact with moving objects is a future direction.
We acknowledge that leveraging LLMs has several problems.
\chatgpt{} may fail when the detector misses the objects and the behavior of \chatgpt{} is regulated by prompts \cite{23icme_actiongpt}.
Despite we have shown in \tabref{tab:object} that \chatgpt{} can be replaced by an open source LLM Mistral 7B, 
using LLMs instead of classical parsing approaches is less efficient in the inference stage.

%% file: sec/7_conclusion.tex
\section{Conclusion}
In this work, we introduce a novel method for generating motion from text in a scene. To tackle this problem, we propose a two-step approach.
The first step involves 3D visual grounding, where we identify the target object. In the second step, concentrating on the target object, we build a diffusion-based motion generation method. Our approach offers several advantages, including improved 3D visual grounding accuracy and motion quality. 